\documentclass[conference]{IEEEtran}
\IEEEoverridecommandlockouts
\usepackage{cite}
\usepackage{amsmath,amssymb,amsfonts}
\usepackage{algorithmic}
\usepackage{graphicx}
\usepackage{textcomp}
\usepackage{xcolor}
\def\BibTeX{{\rm B\kern-.05em{\sc i\kern-.025em b}\kern-.08em
    T\kern-.1667em\lower.7ex\hbox{E}\kern-.125emX}}
\usepackage{xcolor}
\usepackage{datetime}
\usepackage{amsmath}
\usepackage{textcomp}
\usepackage[style=base]{subcaption}
\usepackage{booktabs}
\usepackage{gensymb}
\usepackage{url}

\begin{document}

	\title{
		Autonomous
		Drone Landing with Fiducial Markers and a Gimbal-Mounted Camera for Active Tracking
	}

	\author{
		\IEEEauthorblockN{Joshua Springer}
		\IEEEauthorblockA{\textit{Department of Computer Science} \\
		\textit{Reykjavík University}\\
		Reykjavík, Iceland \\
		\texttt{joshua19@ru.is}}
		\and
		\IEEEauthorblockN{Marcel Kyas}
		\IEEEauthorblockA{\textit{Department of Computer Science} \\
		\textit{Reykjavík University}\\
		Reykjavík, Iceland \\
		\texttt{marcel@ru.is}}
	}

	\maketitle

	\begin{abstract}
		Precision landing is a remaining challenge in autonomous drone flight.
Fiducial markers provide a computationally cheap way for a drone to locate a landing pad and
autonomously execute precision landings.
However, most work in this field depends on either rigidly-mounted or downward-facing cameras which restrict the drone's ability to detect the marker.
We present a method of autonomous landing that uses a gimbal-mounted camera
to quickly search for the landing pad by simply spinning in place while tilting the camera up and down,
and to continually aim the camera at the landing pad during approach and landing.
This method demonstrates successful search, tracking, and landing with 4 of 5 tested fiducial systems on a physical drone with no human intervention.
Per fiducial system, we present
the distributions of the distances from the drone to the center of the landing pad after each successful landing.
We also show representative examples of flight trajectories,
marker tracking performance,
and control outputs for each channel during the landing.
Finally, we discuss qualitative strengths and weaknesses underlying
each system.

	\end{abstract}

	\begin{IEEEkeywords}
		drone, autonomous, fiducial, precision, landing
	\end{IEEEkeywords}

	\section{Introduction}
	\label{section:introduction}
Precision landing is a crucial part of autonomous drone flight
that does not yet have a widespread solution,
although several projects offer potential solutions (see Section~\ref{section:related_work}).
GPS is the main navigational tool of drones in general,
but it does not provide enough positioning accuracy for landing in
e.g.~bad weather, urban canyons, and extreme latitudes.
\textit{Fiducial markers} (see Section~\ref{section:background})
provide a computationally cheap way
for a drone to more accurately estimate its position relative to the landing pad
using just a monocular camera -- arguably the most common drone peripheral sensor.
However, most such projects
use a rigidly-mounted camera,
or a gimbal-mounted (for stabilization) but downward-facing camera,
making it easy for the drone
to lose sight of the landing pad in antagonistic conditions such as wind gusts,
and making it difficult for the drone to search for a landing pad
because it must translate through its environment to see new areas.

We contribute a method of landing with fiducial markers
and a gimbal-mounted camera that actively tracks the landing pad during approach and descent,
and also allows the drone to search for the landing pad
by merely spinning in place and tilting the camera up and down.
However, the actuating camera complicates the system in multiple ways.
First, the tracking system requires extra
components
in order to aim the camera at the gimbal correctly:
the pixel position of the marker in the camera frame, and a controller to aim the gimbal.
Second, the pose estimation from the fiducial marker to the drone is more difficult.
Systems with rigidly mounted cameras can deduce the orientation of the marker
using static transforms from the drone's orientation given by its intertial measurement unit (IMU),
and systems with stabilized, downward-facing cameras can use only the position of the marker.
When the camera is tilting up and down,
we must use either dynamic transforms from the drone to the camera,
or the orientation of the marker itself to determine the drone's pose relative to the landing pad.
The problem here is that many commercially-available gimbals
do not provide their orientation data as an output,
and the orientations of fiducial markers are often ambiguous (see Section~\ref{section:background}).
It is also possible to add an IMU, to the camera to extract its orientation,
but we prefer to require fewer physical components in order to make the system more generalizable.
Instead, we reasonably assume that the landing pad is level,
and transform the pose of the detected fiducial marker using its somewhat unreliable orientation.
As a precursor to this study, we conducted tests of 5 fiducial systems to determine their prevalence of orientation ambiguity~\cite{fiducial_precursor_evaluation},
and developed the landing method in simulation~\cite{joshua_master_thesis}.

	\section{Background}
	\label{section:background}

Fiducial markers
-- such as April Tag~\cite{apriltag3_paper},
WhyCode~\cite{whycode_paper},
ARTag~\cite{ar_tag},
ArUco~\cite{aruco_orig}, etc.
-- are 2D patterns whose pose (position + orientation) can be determined computationally
cheaply from monocular images.
While most fiducial systems provide accurate \textit{position} estimates
their \textit{orientation} estimates are often ambiguous.
This results in sign flips in the Euler components
of perceived orientations,
which propagate through subsequent calculations,
and cause erratic behavior or crashes if a drone's control signal derives from a transformed pose.

\begin{figure}[]
    \centering
    \begin{subfigure}[b]{0.45\linewidth}
        \includegraphics[width=\textwidth]{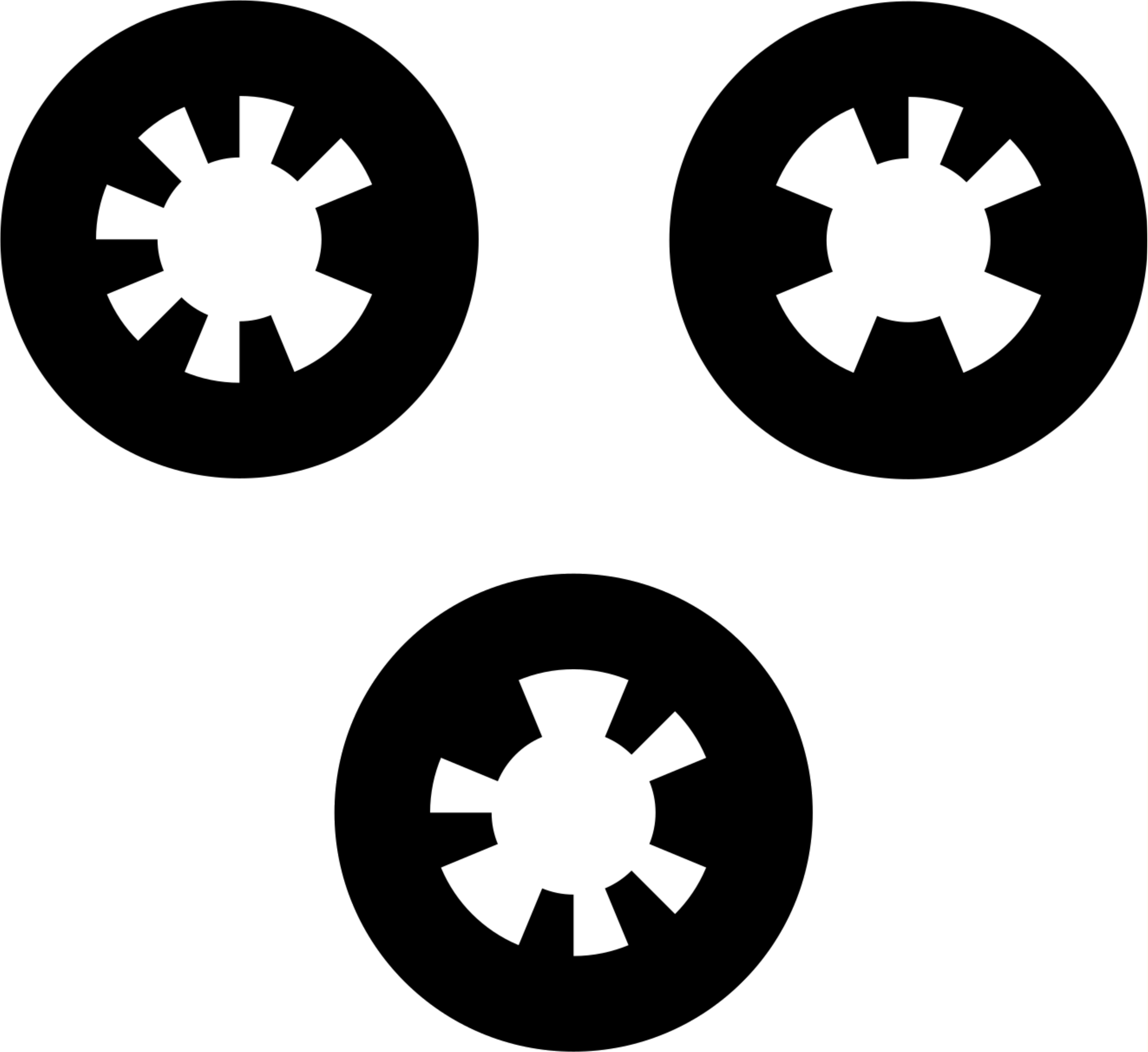}
        \caption{WhyCode ``Bundle''}
        \label{figure:whycode_bundle}
    \end{subfigure}
    \begin{subfigure}[b]{0.45\linewidth}
        \includegraphics[width=\textwidth]{./images/tagCustom24h10_00002_00001_00000.pdf}
        \caption{April Tag 24h10}
        \label{figure:apriltag24h10}
    \end{subfigure}

    \begin{subfigure}[b]{0.45\linewidth}
        \includegraphics[width=\textwidth]{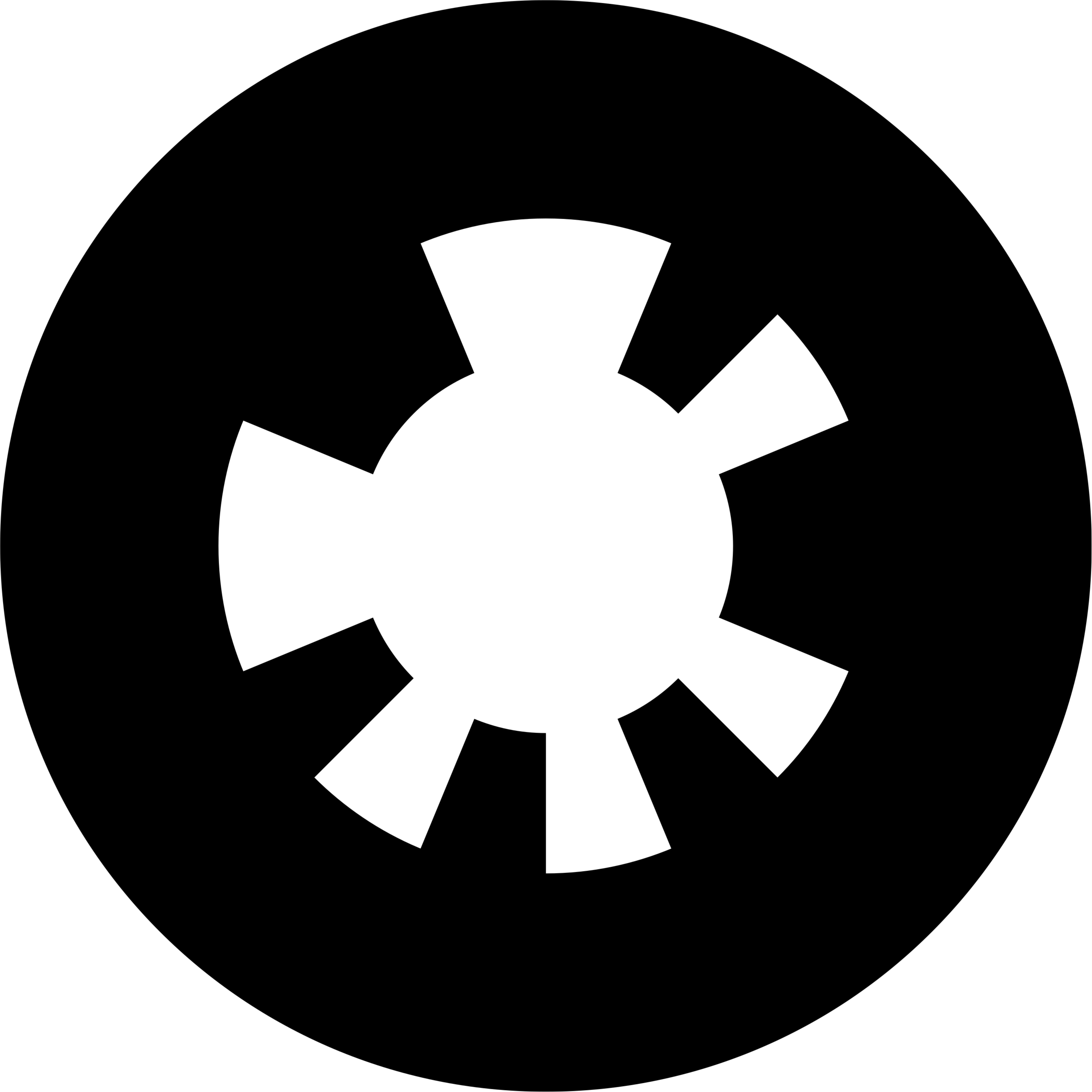}
        \caption{WhyCode}
        \label{figure:whycode_single}
    \end{subfigure}
    \begin{subfigure}[b]{0.45\linewidth}
        \includegraphics[width=\textwidth]{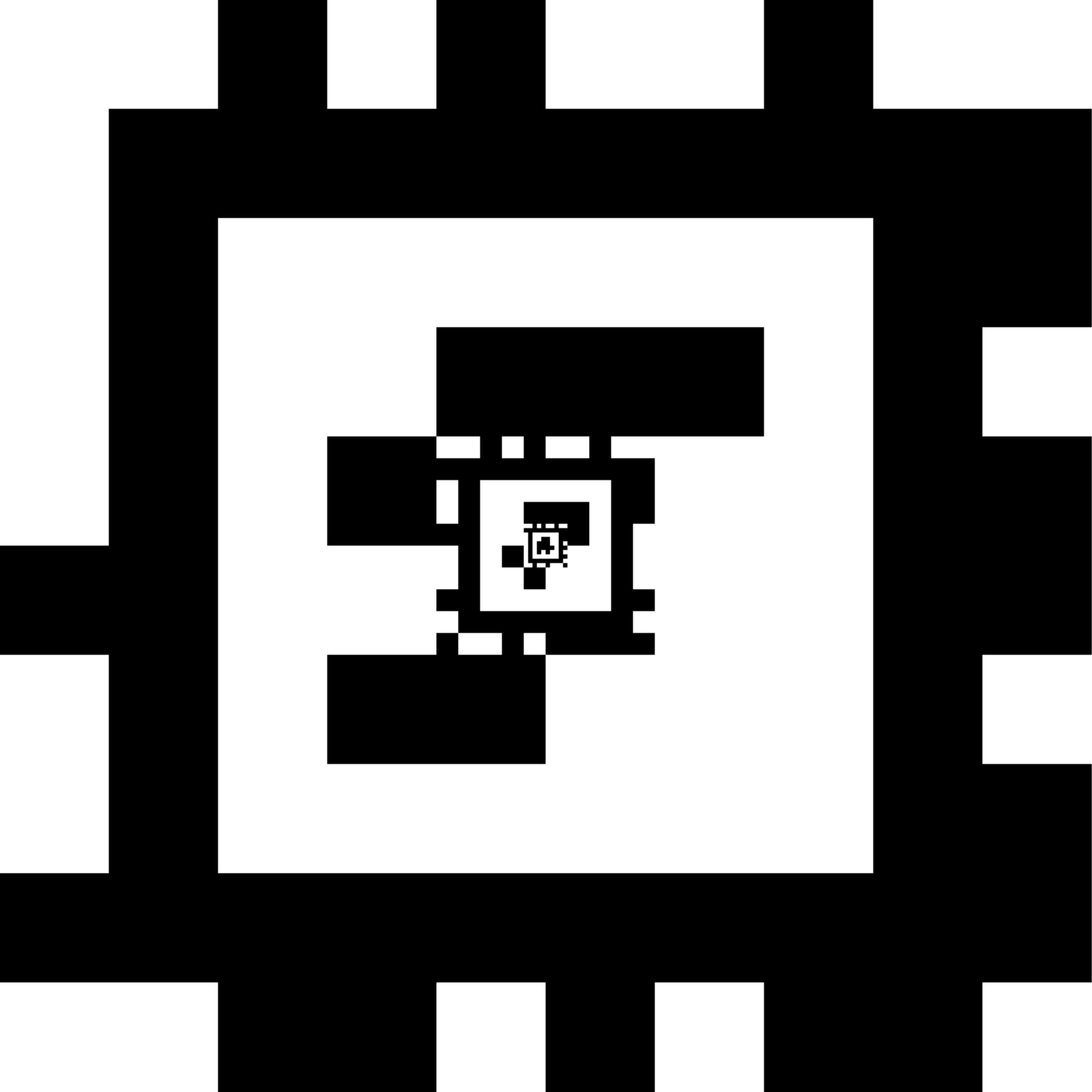}
        \caption{April Tag 48h12}
        \label{figure:apriltag48h12}
    \end{subfigure}
    \caption{The 4 landing pads in this paper.}
    \label{figure:marker_setup}
\end{figure}

In a precursor study~\cite{fiducial_precursor_evaluation},
we evaluate 2 existing fiducial systems
-- April Tag 48h12 (Figure~\ref{figure:apriltag48h12}) and WhyCode (Figure~\ref{figure:whycode_single})
-- and 3 modified versions of those systems
-- April Tag 24h10 (Figure~\ref{figure:apriltag24h10}),
WhyCode Ellipse (Figure~\ref{figure:whycode_single}),
and WhyCode Multi (Figure~\ref{figure:whycode_bundle})
-- in terms of orientation ambiguity and runtime framerate when executing on a Raspberry Pi 4 (2GB RAM).
WhyCode Ellipse changes the sampling locations for the decision problem that determines
the orientation of WhyCode markers,
WhyCode Multi uses the positions of multiple coplanar markers to determine
the orientation of the plane connecting them (and then assumes they all have this orientation),
and April Tag 24h10 has a layout that is comparable to but smaller than April Tag 48h12,
in the hopes of more computational cheapness.
We conclude that, in the context of our testing environment, the systems can be ranked
in order of increasing prevalence of orientation ambiguity:
WhyCode Ellipse, April Tag 48h12, WhyCode Orig, WhyCode Multi, and April Tag 24h10;
and in order of decreasing runtime detection rate:
WhyCode Orig, WhyCode Ellipse, WhyCode Multi, April Tag 48h12, and April Tag 24h10.

	\section{Related Work}
	\label{section:related_work}

Some projects have accomplished precision drone landing with fiducial markers
and a camera that is downward-facing and rigidly-mounted,
or gimbal-mounted (for stabilization) but oriented vertically downwards.
In some cases~\cite{wynn,accurate_landing_UAV_ground_pattern},
they use multiple ArUco markers of different sizes
to maintain detection of the landing pad as the drone approaches very close.
In the case of~\cite{vision_based_x_platform},
the drone detects a single, custom X-shaped marker with 2 fixed cameras
-- one pointing down and the other pointing forward and down.
In~\cite{fiducial_vessel_landing_ar_tag_two_fixed_cameras,fiducial_landing_two_fixed_cameras_apriltag},
the drone has a front-facing camera and a downward facing camera,
and detects a single marker (AR Tag, April Tag respectively) for landing on a moving boat.
One project~\cite{high_velocity_landing}
uses GPS for an initial approach, and an April Tag for final approach.
One method~\cite{fiducial_landing_downward_facing_90_deg_gimbaled_camera} uses a DJI Phantom 4,
which has a gimbal-mounted camera,
but keeps the gimbal pointed straight down at the ground when detecting fiducial markers.
The drone in~\cite{fiducial_landing_many_markers_voting_fixed_camera} uses a fixed,
downward facing camera,
but does address the issue of orientation ambiguity using many co-planar April Tag markers,
to determine the position of the landing pad through voting.
In~\cite{fiducial_landing_ship_6dof_single_fixed_downfront_camera_apriltag},
the drone has a single camera facing forwards and down,
and detects April Tag markers on its landing pad.
Finally, one method~\cite{lentimark_landing} uses a closed-source,
marker system called Lentimark~\cite{lentimark},
which mitigates the orientation ambiguity problem
using Moir\'e patterns on the outside of otherwise conventional AR Tags.
They embed a single Lentimark marker inside of an AR Tag,
mount them on a post, and land a drone autonomously
while tracking the marker during landing with a gimbal-mounted camera.
The drone lands on the ground in front of the post.

Some of these projects report visual loss of the marker during landing,
as a result of the lack of camera tracking.
We contribute a method that goes beyond these projects to actively track the marked landing pad
with a gimbal-mounted camera independently of the drone's movement,
enabling both autonomous precision landing in unfriendly environments (e.g.~wind)
and the ability to
safely search for the marker
by simply spinning in place and tilting the camera up and down.
The drone lands on
the marked landing pad.
We focus on open source marker systems and also test some systems that have not appeared in
autonomous landing scenarios.


	\section{Methods}
	\label{section:methods}


\emph{System overview:} Our testing platform is a DJI Spark,
which provides the ability to test indoors
(reducing logistical considerations),
and poses little risk in the case of crashes or malfunctions,
It uses the DJI Mobile SDK for autonomous control,
and we have modified DJI's code samples to create a custom app~\cite{our_android_app}
that offloads video frames to a companion board -- a Raspberry Pi 4 (2 GB of RAM).
The companion board's software detects the landing pads in the frames,
generates control commands,
and sends them to the drone via the app and controller.
The control system stays the same, but the landing pad and fiducial software change,
to test each of the 5 fiducial systems in the precursor study (see Section~\ref{section:background}).
Each of the fiducial systems provides exactly the same attributes to the control system:
the relative position from the drone to the landing pad (a \textit{position target}),
the pixel position $u,v$ of the landing pad in the camera frame
(normalized so $u,v \in \left[-1, 1\right]$ for ease of use, where (0,0) is the center),
and the yaw of the landing pad (for alignment in the later stages of landing).
The control system passes these attributes to the DJI Mobile SDK as VirtualStick inputs,
such that they appear to the system as input from the controller,
and the flight controller uses them as velocity setpoints.
While VirtualStick commands can be in the interval $[-1, 1]$,
        each control output is constrained to the interval $[-0.2, 0.2]$
        in order to ensure slow, stable movement in the presence of high latency.
        The throttle is in $[-0.2, 0]$ so it can only command the drone to go down, not up.
        Each control output corresponds to a particular \textit{rate},
        i.e.~yaw corresponds to clockwise yaw velocity,
        gimbal tilt corresponds to the tilt velocity of the gimbal,
        pitch corresponds to the forward velocity,
        roll corresponds to the right velocity,
        and throttle corresponds to the up velocity.

The control policy has several phases.
During the \textit{takeoff} phase, the drone ascends in place to an altitude of 1.2 meters (fully automated within DJI Mobile SDK).
It then transitions to \textit{search}, where it spins in place while tilting the camera up and down.
After it finds the marker, it can transition to \textit{approach}, where it moves toward the landing pad quickly,
without changing altitude.
This phase also has a parameter called ``deadzone'' that disables planar movement if the drone is within a small planar
distance to the marker.
Then, it enters the \textit{yaw align} phase, spinning in place to align with the yaw of the landing pad.
Next, it enters \textit{descent} and decreases altitude until a specific minimum altitude (different for each marker, see Section~\ref{section:results}), while correcting its horizontal position above the landing pad.
It then enters a \textit{landing commit} phase,
where the DJI Mobile SDK handles descent, touchdown detection, and motor shutoff.
Finally, the drone has \textit{landed}.

At the start of each landing attempt, the drone and landing pad are placed 2.5 meters from each other,
with the drone facing directly \textit{away} from the landing pad so that it must search for the landing pad after takeoff.
At the first landing attempt, the landing pad faces away from the drone, and it is rotated clockwise $18 \degree$
after each of the $20$ landings, in order to simulate approaches from all directions.
We consider a landing attempt successful if it requires no human intervention except initialization,
and if the drone touches down fully on the landing pad without touching the ground.


\emph{Basis of Comparison:} We test the landing system using each of the 5 fiducial systems, keeping all other factors the same.
We compare the fiducial systems on the number of successful landings they produce, and on the \textit{accuracy} of the landing -
defined as the ability of the system to minimize the distance from the camera to the center of the marker at touchdown.
All markers have the same width at the widest point, but different areas because they are different shapes.

	\section{Results}
	\label{section:results}

\begin{figure}[]
    \centering
    \includegraphics[width=0.9\linewidth]{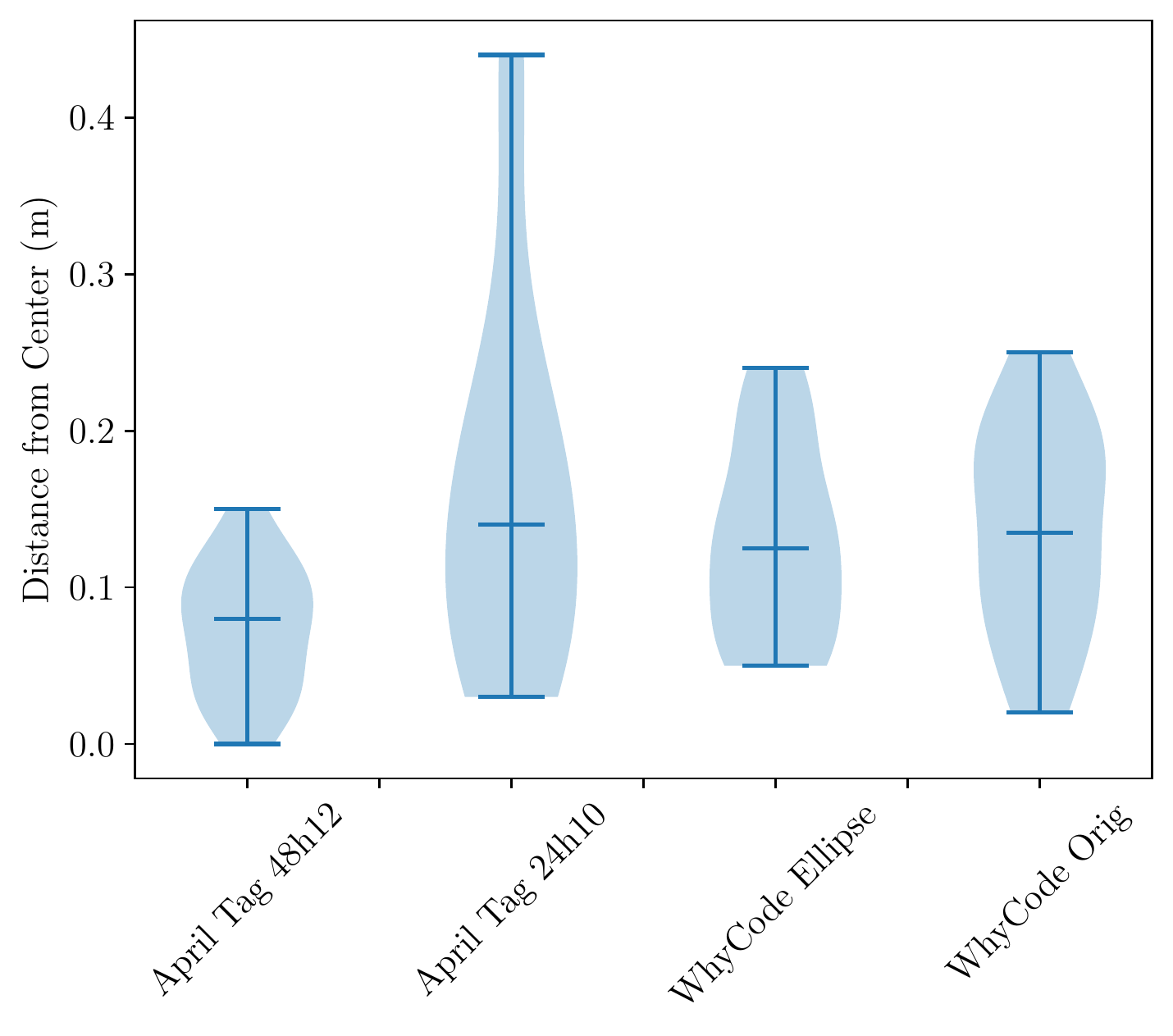}
    \caption{Distances from the camera to the center of the landing pad after touchdown
	for each successful system (which excludes WhyCode Multi). Smaller distances are more accurate. }
    \label{figure:landing_radii}
\end{figure}

\begin{figure}[]
    \centering
    \includegraphics[width=\linewidth]{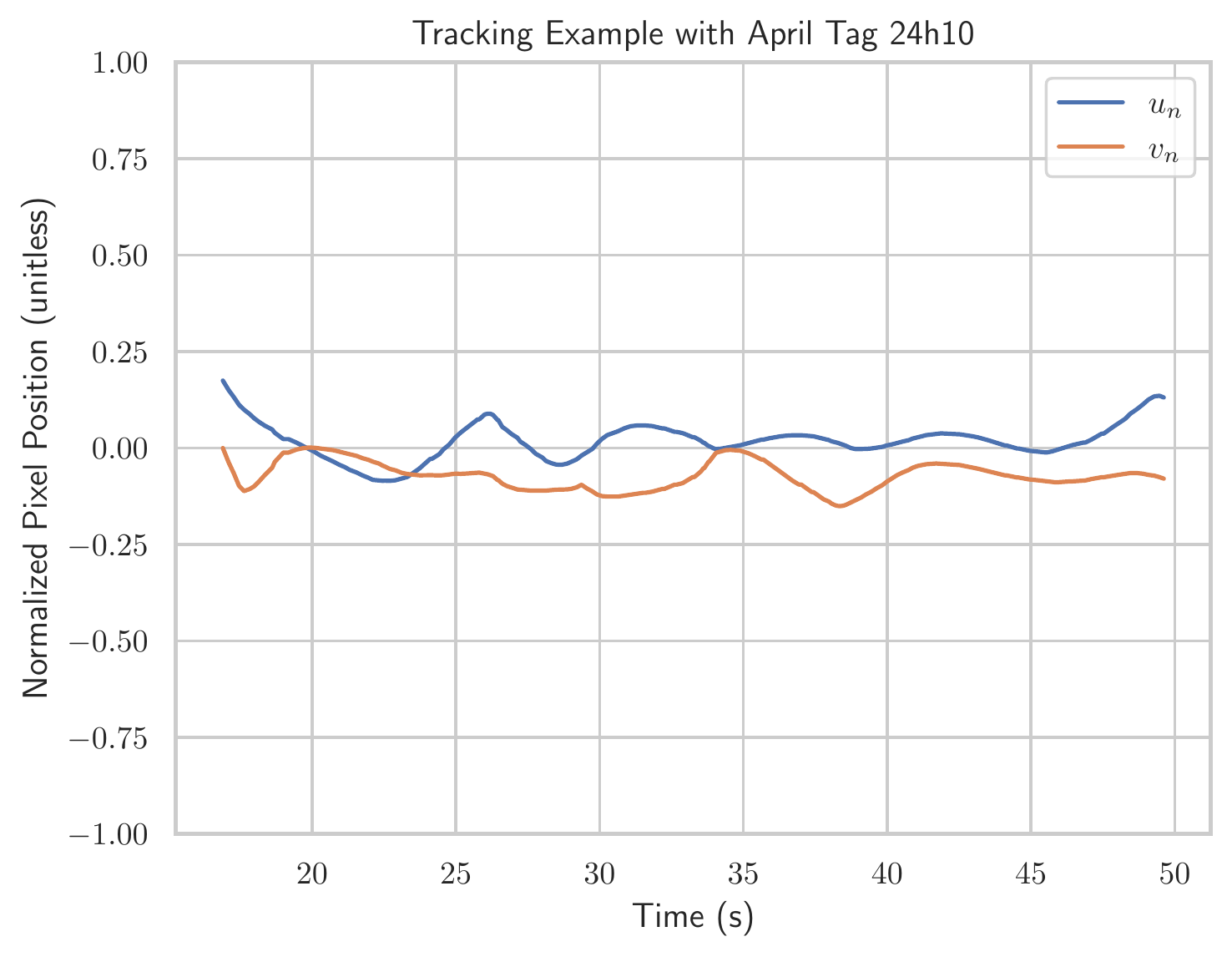}
    \caption
    {
        Normalized pixel positions of the landing pad during an example landing.
        The drone is able to track the marker (i.e.~$(u_n,v_n) \approx (0,0)$)
	by aiming the camera as it approaches.
    }
    \label{figure:example_tracking}
\end{figure}

\begin{figure}[]
    \centering
    \includegraphics[width=\linewidth]{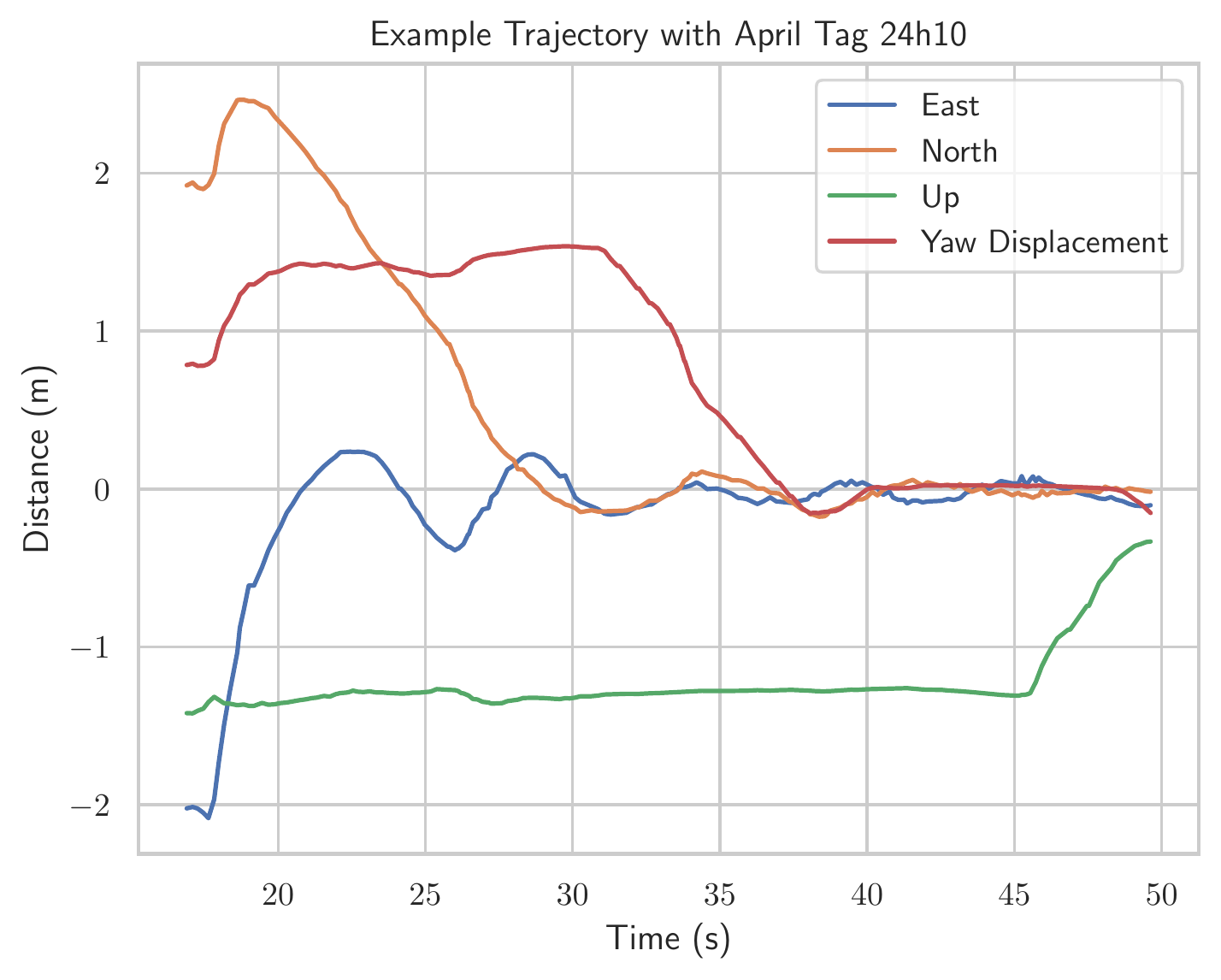}
    \caption
    {
	An example landing trajectory \textit{without} discontinuities.
	The East (left/right) position target decreases as the drone turns to face the landing pad.
	The North (forward/backward) position target decreases slowly as the drone approaches the landing pad.
	The drone then rotates to align with the landing pad, reducing its yaw displacement.
	Finally, the Up position target approaches 0 as the drone descends.
    }
    \label{figure:example_approach}
\end{figure}

\begin{figure}[]
    \centering
    \includegraphics[width=\linewidth]{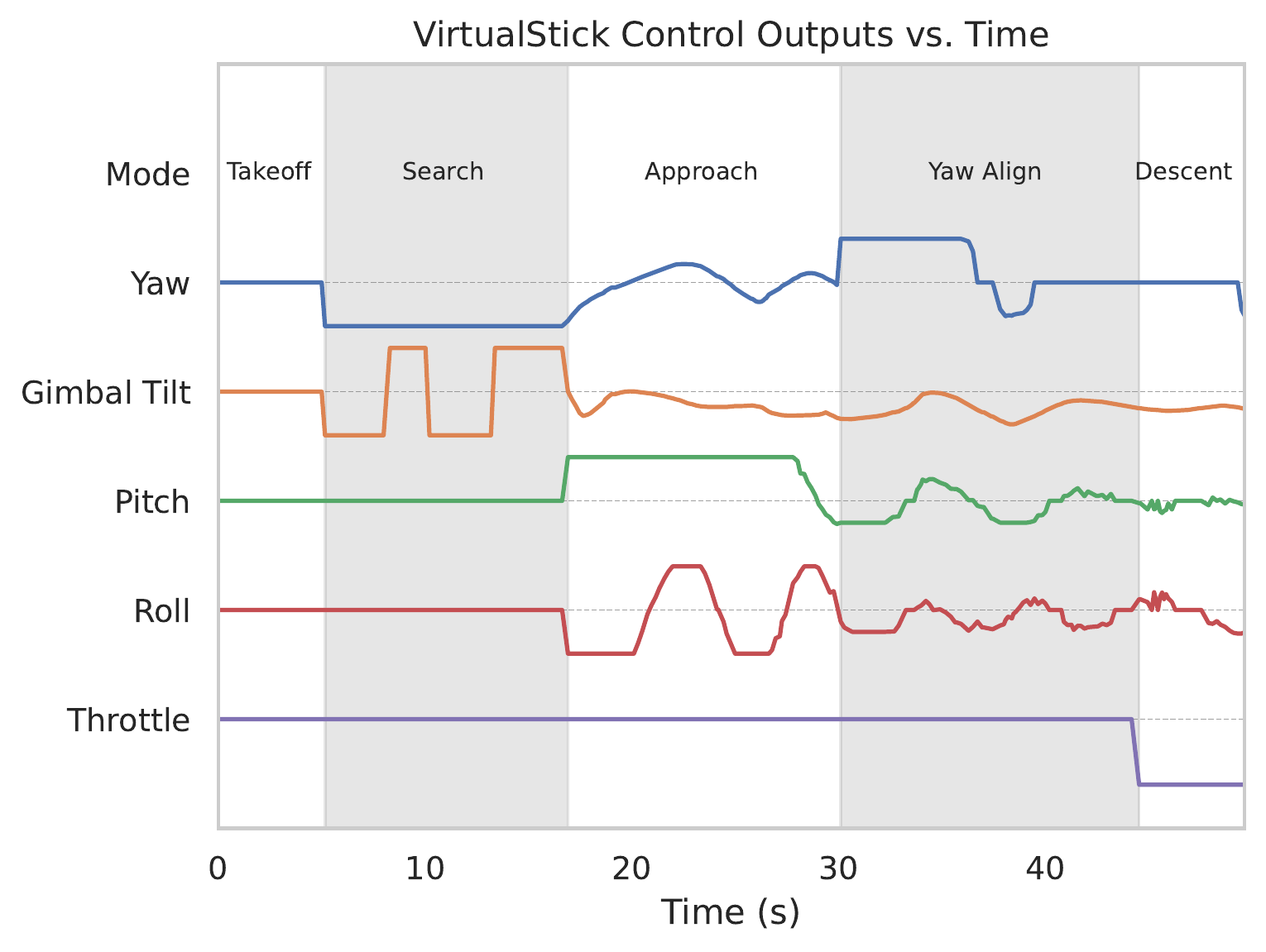}
    \caption
    {
        VirtualStick commands sent to the drone during an example landing.
        \textit{Takeoff:} control outputs are neutral.
        \textit{Search:} the drone rotates counterclockwise, and the gimbal tilts up and down.
        \textit{Approach:} the drone tracks the landing pad with yaw and gimbal tilt, approaching with pitch and roll.
        \textit{Yaw align:} the drone tracks the landing pad with gimbal tilt, maintains its position, and aligns to the landing pad's yaw.
        \textit{Descent:} the drone maintains its horizontal position and descends.
    }
    \label{figure:example_control}
\end{figure}

\begin{figure}[]
    \centering
    \includegraphics[width=\linewidth]{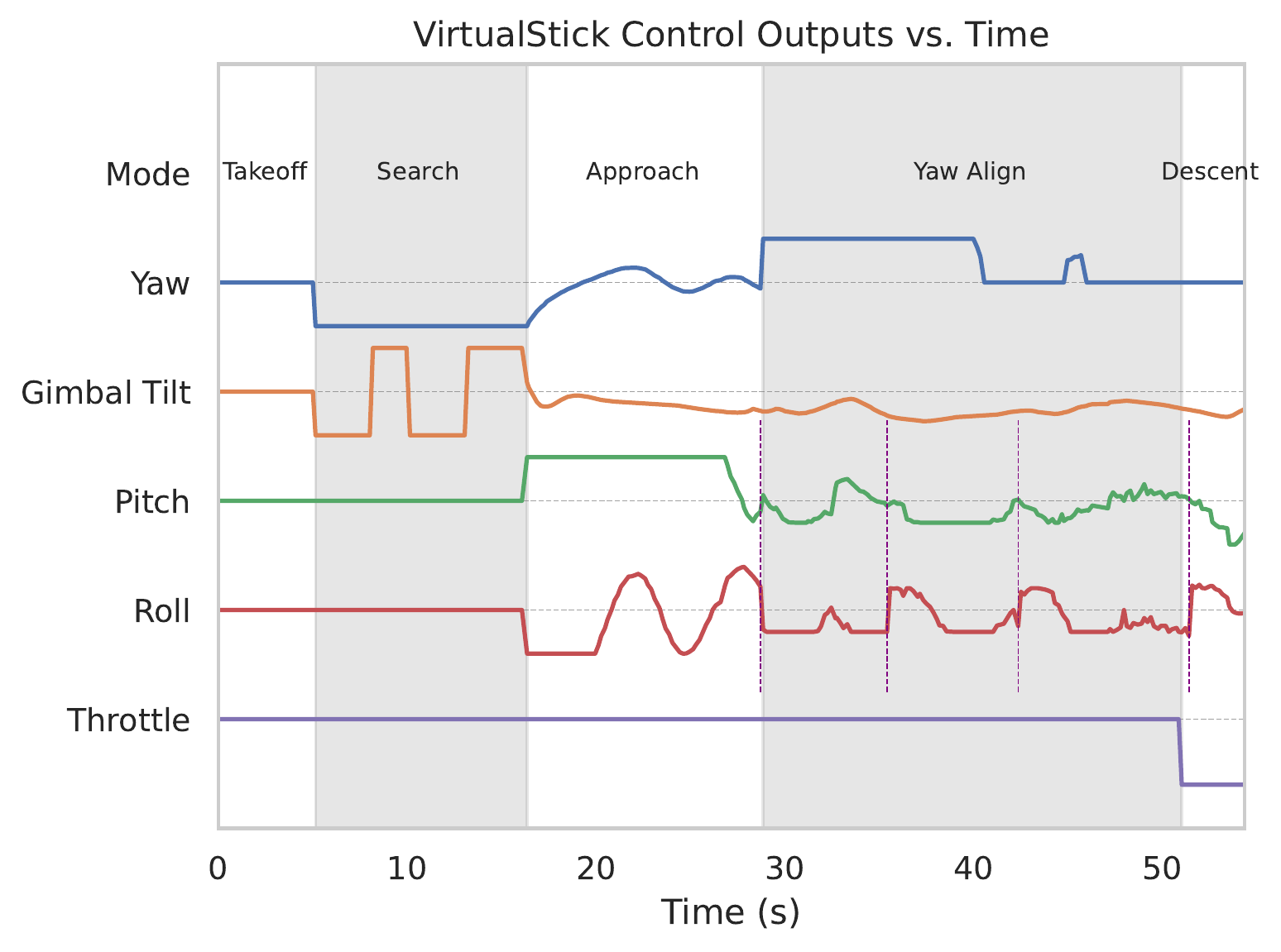}
    \caption
    {
	    The VirtualStick commands
	during a
	landing
	with discontinuities (shown by vertical lines)
	that cause the drone to persistently oscillate around the landing pad.
    }
    \label{figure:example_control_with_discontinuities}
\end{figure}

All systems achieved 20 successful landings, except for WhyCode Multi, which achieved none.
For each successful system,
Figure~\ref{figure:landing_radii} shows the distribution of the distances
from the camera to the center of the landing pad after touchdown,
which is ideally 0.
Lower values represent more accurate landings.
For one representative landing that is not affected by orientation ambiguity,
Figure~\ref{figure:example_approach} shows the perceived positions
of the drone during its approach
Figure~\ref{figure:example_tracking} shows the pixel positions of the landing pad
and Figure~\ref{figure:example_control} shows the control outputs.
Figure~\ref{figure:example_control_with_discontinuities} shows the control outputs for a landing
that is affected by orientation ambiguity.



%


\begin{figure}[]
    \centering
    \includegraphics[width=\linewidth]{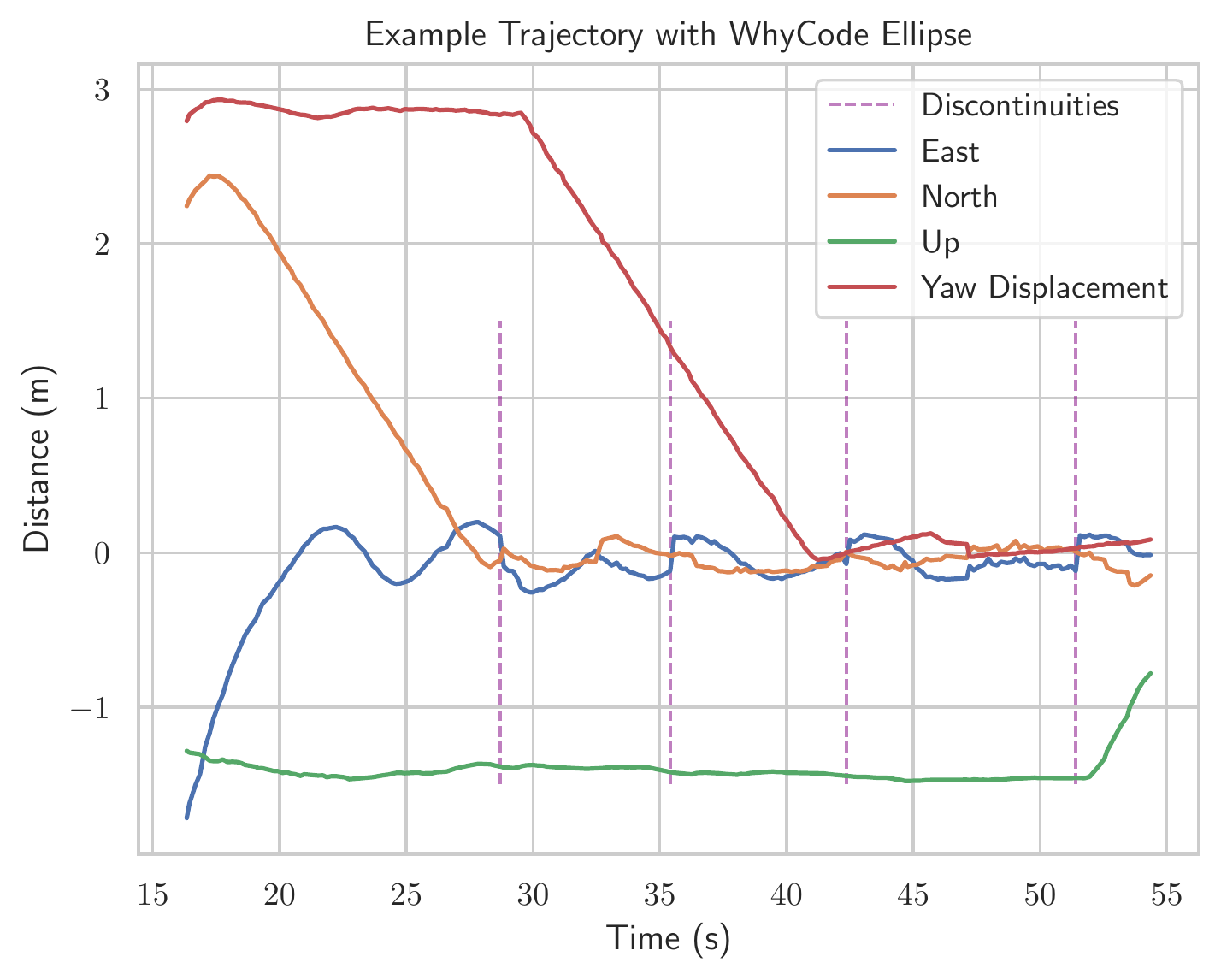}
    \caption
    {
        An example landing trajectory with discontinuities,
	shown by vertical lines at $t\in\{28.7, 35.4, 42.4, 51.4\}$ (s), and they propagate to the control signals shown in Figure~\ref{figure:example_control_with_discontinuities}.
    }
    \label{figure:example_approach_with_discontinuities}
\end{figure}

	\section{Discussion}
	\label{section:discussion}


\emph{Fiducial System Performance:} April Tag 48h12 provides the most accurate landings,
while April Tag 24h10, WhyCode Ellipse, and WhyCode Orig have similar levels of accuracy to each other.
April Tag 24h10 generates a lot of erronous control signals,
but its concentric marker structure allows it to track the landing pad at low altitudes ($< 0.4$ meters).
WhyCode Ellipse and WhyCode Orig exhibit less orientation ambiguity
but require higher \emph{landing commit} heights,
below which the rest of their landing is blind.

WhyCode Multi unfortunately did not demonstrate any successful landings,
although it did provide good performance in the \textit{search} phase and the initial part of the \textit{approach} phase.
However, it invariably overshot the landing pad every time.
While it outputs correct position targets in lab experiments,
its position targets are, apparently,
highly dependent on the angle from which the marker is viewed.
The camera for the spark can only point between $0\degree$ (straight forward) and $-85\degree$ down.
It is possible that the drone could recover from an overshoot
if the gimbal were able to point backwards at the marker.

The prevelance of orientation ambiguity
correlates to the distance from the center of the landing pad to the drone's camera after landing,
because of the resulting, erroneous control signals.
Further, when the drone is almost directly above the landing pad,
orientation ambiguity is high
and can cause the drone to move from one side of the landing pad to the other
in stable oscillation -- hence the deadzone.

All systems
occasionally lost sight
of the landing pad,
contributing to a small number of landing failures.
If this happened in a mission scenario, the system could simply begin searching for the landing pad again.
Visual loss of the April Tags typically occurred upon initial acquisition of the landing pad,
when they were harder to identify as a result of their
small pixel area (because of the long distance) and motion blur
(because of the panning and tilting camera).
After initial identification,
however, they provide reliable detection even at very close distances (because of their concentric marker embedding).
WhyCode markers exhibit reliable long-distance detection even under motion,
but eclipse the camera's field of view completely at close range.
Finally, some visual loss occurs simply as a result of sun glares or shadows
-- an inherent problem in identifying markers in the visible spectrum.

Perhaps the most interesting result of this study is that
the orientation ambiguity problem does not prohibit autonomous
precision landing using a gimbal-mounted camera that is tracking the marker.
Orientation ambiguity caused discontinuities in the control outputs,
but they caused only minor disturbances
-- not destructive interference.
Importantly, most of the discontinuities occur when the drone is almost directly above the landing pad,
since the orientation is hardest to perceive when the camera is normal to the fiducial marker.
In these cases, the drone's planar radius to the landing pad is small,
so it is likely inside the deadzone
where the erroneous signals have no effect on the drone's behavior.


\emph{Drone Platform Difficulties:}
The DJI Mobile SDK connects to an app on a tablet
for
displaying video, and receiving user input.
However, the tablet is not representative of the hardware that will
eventually be embedded onboard a drone,
so, the app must offload video frames to a Raspberry Pi 4 via a WiFi connection
(no more wired interfaces were available).
This transmission
required converting the frames from
color \texttt{.bmp}
to
grayscale \texttt{.webp} at 20\% quality to achieve an adequately fast system
with a maximum framerate of about 7 Hz,
and caused an inconsistent latency of between 0.5 and 2 seconds
from image acquisition to control output.
The low framerate and inconsistent processing times
left artifacts in the experiments,
such as the oscillations in the roll signal
of Figure~\ref{figure:example_control}.
As the drone approaches the landing pad head-on, it over-corrects for left-right positional offset
because it does not receive the video stream quickly enough.


	\section{Conclusion \& Future Work}
	\label{section:conclusion}

We have shown that autonomous precision landing is possible with fiducial markers and a gimbal-mounted camera for active tracking.
Even with high processing latency,
4 of 5 tested systems have demonstrated success in the real world.
While our platform does not allow for on onboard companion board,
all essential image processing and control signal generation
is occurs in real time onboard a Raspberry Pi,
meaning that the entire system can be embedded onto a larger drone.
We have also demonstrated that
the gimbal-mounted camera setup allows the drone to search for the landing pad by simply
spinning in place and sweeping its gimbal up and down,
thereby safely and efficiently scanning a large area.
\label{section:future_work}

The next step is to re-conduct the experiments with the following changes:
embed the companion board into a larger drone to reduce the system's latency,
Use a gimbal that is able to point backwards to potentially improve WhyCode Multi's performance.
Extensions of this work should explore methods for filtering erroneous control signals
that result from orientation ambiguity,
or use an IMU on the drone's camera
to remove the problem of orientation ambiguity altogether.

	%

	\bibliographystyle{plain}
	\bibliography{main}

\begin{thebibliography}{10}

\bibitem{fiducial_landing_many_markers_voting_fixed_camera}
{Araar, Oualid and Aouf, Nabil and Vitanov, Ivan}.
\newblock {Vision Based Autonomous Landing of Multirotor UAV on Moving
  Platform}.
\newblock {\em {Journal of Intelligent \& Robotic Systems}}, 85, 02 2017.

\bibitem{high_velocity_landing}
Alexandre Borowczyk, Duc-Tien Nguyen, André {Phu-Van Nguyen}, Dang~Quang
  Nguyen, David Saussié, and Jerome~Le Ny.
\newblock {Autonomous Landing of a Multirotor Micro Air Vehicle on a High
  Velocity Ground Vehicle}.
\newblock {\em IFAC-PapersOnLine}, 50(1):10488--10494, 2017.
\newblock 20th IFAC World Congress.

\bibitem{fiducial_landing_ship_6dof_single_fixed_downfront_camera_apriltag}
{Chaves, Stephen and Wolcott, Ryan and Eustice, Ryan}.
\newblock {NEEC Research: Toward GPS-denied Landing of Unmanned Aerial Vehicles
  on Ships at Sea}.
\newblock {\em {Naval Engineers Journal}}, 127:23--35, 03 2015.

\bibitem{vision_based_x_platform}
Davide Falanga, Alessio Zanchettin, Alessandro Simovic, Jeffrey Delmerico, and
  Davide Scaramuzza.
\newblock {Vision-based Autonomous Quadrotor Landing on a Moving Platform}.
\newblock 10 2017.

\bibitem{ar_tag}
Mark Fiala and Mark Fiala.
\newblock {ARTag, a Fiducial Marker System Using Digital Techniques}.
\newblock In {\em Proceedings of the 2005 IEEE Computer Society Conference on
  Computer Vision and Pattern Recognition (CVPR'05) - Volume 2 - Volume 02},
  CVPR '05, pages 590--596, Washington, DC, USA, 2005. IEEE Computer Society.

\bibitem{aruco_orig}
Sergio Garrido-Jurado, Rafael Muñoz-Salinas, Francisco Madrid-Cuevas, and
  Manuel Marín-Jiménez.
\newblock {Automatic generation and detection of highly reliable fiducial
  markers under occlusion}.
\newblock {\em Pattern Recognition}, 47:2280–2292, 06 2014.

\bibitem{our_android_app}
{Joshua Springer}.
\newblock {Edited Video Decoding Sample (for Autonomous Landing)}.
\newblock \url{https://github.com/uzgit/Android-VideoStreamDecodingSample},
  2021.

\bibitem{fiducial_precursor_evaluation}
{Joshua Springer and Marcel Kyas}.
\newblock {Evaluation of Orientation Ambiguity and Detection Rate in April Tag
  and WhyCode}.
\newblock In {\em {2022 IEEE International Conference on Robotic Computing}},
  December 2022.

\bibitem{apriltag3_paper}
M.~{Krogius}, A.~{Haggenmiller}, and E.~{Olson}.
\newblock Flexible layouts for fiducial tags.
\newblock In {\em 2019 IEEE/RSJ International Conference on Intelligent Robots
  and Systems (IROS)}, pages 1898--1903, 2019.

\bibitem{fiducial_landing_two_fixed_cameras_apriltag}
Tianpei Liao, Amaldev Haridevan, Yibo Liu, and Jinjun Shan.
\newblock {Autonomous Vision-based UAV Landing with Collision Avoidance using
  Deep Learning}, 2021.

\bibitem{whycode_paper}
Peter Lightbody, Tom\'{a}\v{s} Krajn\'{\i}k, and Marc Hanheide.
\newblock {A Versatile High-performance Visual Fiducial Marker Detection System
  with Scalable Identity Encoding}.
\newblock In {\em Proceedings of the Symposium on Applied Computing}, SAC '17,
  pages 276--282, New York, NY, USA, 2017. ACM.

\bibitem{fiducial_landing_downward_facing_90_deg_gimbaled_camera}
Phong~Ha Nguyen, Ki~Wan Kim, Young~Won Lee, and Kang~Ryoung Park.
\newblock {Remote Marker-Based Tracking for UAV Landing Using Visible-Light
  Camera Sensor}.
\newblock {\em Sensors}, 17(9), 2017.

\bibitem{fiducial_vessel_landing_ar_tag_two_fixed_cameras}
{Polvara, Riccardo and Sharma, Sanjay and Wan, Jian and Manning, Andrew and
  Sutton, Robert}.
\newblock {Vision-Based Autonomous Landing of a Quadrotor on the Perturbed Deck
  of an Unmanned Surface Vehicle}.
\newblock {\em Drones}, 2(2), 2018.

\bibitem{joshua_master_thesis}
Joshua Springer.
\newblock {Autonomous Landing of a Multicopter Using Computer Vision}.
\newblock Master's thesis, Reykjavík University, 2020.
\newblock \url{http://hdl.handle.net/1946/36422}.

\bibitem{lentimark_landing}
Hideyuki Tanaka and Yoshio Matsumoto.
\newblock Autonomous drone guidance and landing system using ar/high-accuracy
  hybrid markers.
\newblock In {\em 2019 IEEE 8th Global Conference on Consumer Electronics
  (GCCE)}, pages 598--599, 2019.

\bibitem{lentimark}
Hideyuki Tanaka, Kunihiro Ogata, and Yoshio Matsumoto.
\newblock Solving pose ambiguity of planar visual marker by wavelike two-tone
  patterns.
\newblock In {\em 2017 IEEE/RSJ International Conference on Intelligent Robots
  and Systems (IROS)}, pages 568--573, 2017.

\bibitem{accurate_landing_UAV_ground_pattern}
Jamie Wubben, Francisco Fabra, Carlos Calafate, Tomasz Krzeszowski, Johann
  Marquez-Barja, Juan-Carlos Cano, and Pietro Manzoni.
\newblock {Accurate Landing of Unmanned Aerial Vehicles Using Ground Pattern
  Recognition}.
\newblock {\em Electronics}, 8:1532, 12 2019.

\bibitem{wynn}
Jesse~S. Wynn.
\newblock {Visual Servoing for Precision Shipboard Landing of an Autonomous
  Multirotor Aircraft System}.
\newblock Master's thesis, Brigham Young University, 9 2018.

\end{thebibliography}

\end{document}